\documentclass[10pt,twocolumn,letterpaper]{article}

\usepackage{iccv}
\usepackage{times}
\usepackage{epsfig}
\usepackage{graphicx}
\usepackage{amsmath}
\usepackage{amssymb}

\usepackage{multirow}
\usepackage{booktabs}
\usepackage{verbatim}
\usepackage{paralist}
\usepackage{subcaption}

\newcommand\norm[1]{\left\lVert#1\right\rVert}

\usepackage[accsupp]{axessibility}  % Improves PDF readability for those with disabilities.

\usepackage[breaklinks=true,bookmarks=false]{hyperref}

\iccvfinalcopy % *** Uncomment this line for the final submission

\def\iccvPaperID{4826} % *** Enter the ICCV Paper ID here

\ificcvfinal\fi

\begin{document}

%%%%%%%%% TITLE
\title{ExBluRF: Efficient Radiance Fields for \\ Extreme Motion Blurred Images}

\author{
  Dongwoo Lee$^{1}$ \hskip1.6em Jeongtaek Oh$^{2}$ \hskip1.6em Jaesung Rim$^{3}$ \hskip1.6em Sunghyun Cho$^{3,4}$ \hskip1.6em Kyoung Mu Lee$^{1,2}$ \\
   $^{1}$Dept. of ECE\&ASRI, $^{2}$IPAI, Seoul National University, Korea  \\ 
   $^{3}$GSAI, $^{4}$Dept. of CSE, POSTECH, Korea \\ % Graduate School of AI, POSTECH, Korea
   {\tt\small \{dongwoo.lee,ohjtgood\}@snu.ac.kr \hskip1.6em \{jsrim123,s.cho\}@postech.ac.kr \hskip1.6em kyungmu@snu.ac.kr} 
}

\maketitle
\ificcvfinal\thispagestyle{empty}\fi

%%%%%%%%% ABSTRACT
\begin{abstract}
We present ExBluRF, a novel view synthesis method for extreme motion blurred images based on efficient radiance fields optimization.
Our approach consists of two main components: 6-DOF camera trajectory-based motion blur formulation and voxel-based radiance fields.
From extremely blurred images, we optimize the sharp radiance fields by jointly estimating the camera trajectories that generate the blurry images.
In training, multiple rays along the camera trajectory are accumulated to reconstruct single blurry color, which is equivalent to the physical motion blur operation.
We minimize the photo-consistency loss on blurred image space and obtain the sharp radiance fields with camera trajectories that explain the blur of all images.
The joint optimization on the blurred image space demands painfully increasing computation and resources proportional to the blur size.
Our method solves this problem by replacing the MLP-based framework to low-dimensional 6-DOF camera poses and voxel-based radiance fields.
Compared with the existing works, our approach restores much sharper 3D scenes from challenging motion blurred views with the order of $10\times$ less training time and GPU memory consumption.
\end{abstract}

%%%%%%%%% BODY TEXT
\section{Introduction}
Neural Radiance Fields (NeRF) have made great progress on novel view synthesis in recent years.
A number of follow-up works pay attention to NeRF's~\cite{mildenhall2020nerf} photo-realistic view synthesis performance, and focus on improving training~\cite{deng2022depth,tancik2021learned} and rendering~\cite{garbin2021fastnerf,reiser2021kilonerf} speed for practical applications.
However, enhancing NeRF's rendering quality from degraded multi-view images is yet to be explored extensively.

Camera motion blur is a representative degradation of images taken under low-light conditions and camera shake.
Optimizing NeRF from the motion blurred images suffers from the severe shape-radiance ambiguity~\cite{zhang2020nerf++} and produces inaccurate 3D geometry reconstruction and low-quality view synthesis of the scene.

\begin{figure}[t]
	\vspace{-0.1in}
	\centering
	\subfloat[Blurry Views]{\includegraphics[height=5.58cm]{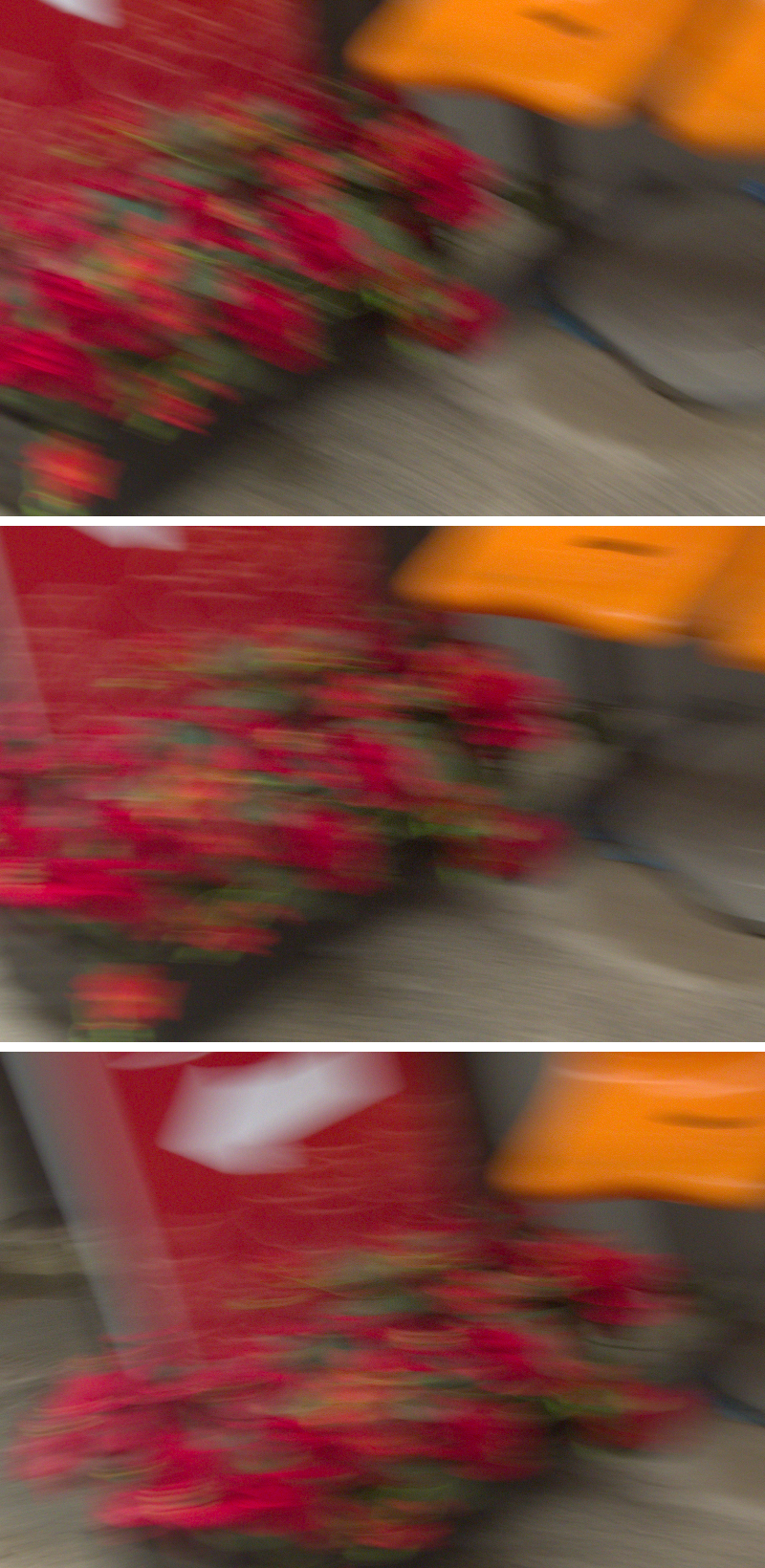}}\hspace*{-0.05in}
        \hfill
	\subfloat[Rendered Novel View]{\includegraphics[height=5.58cm]{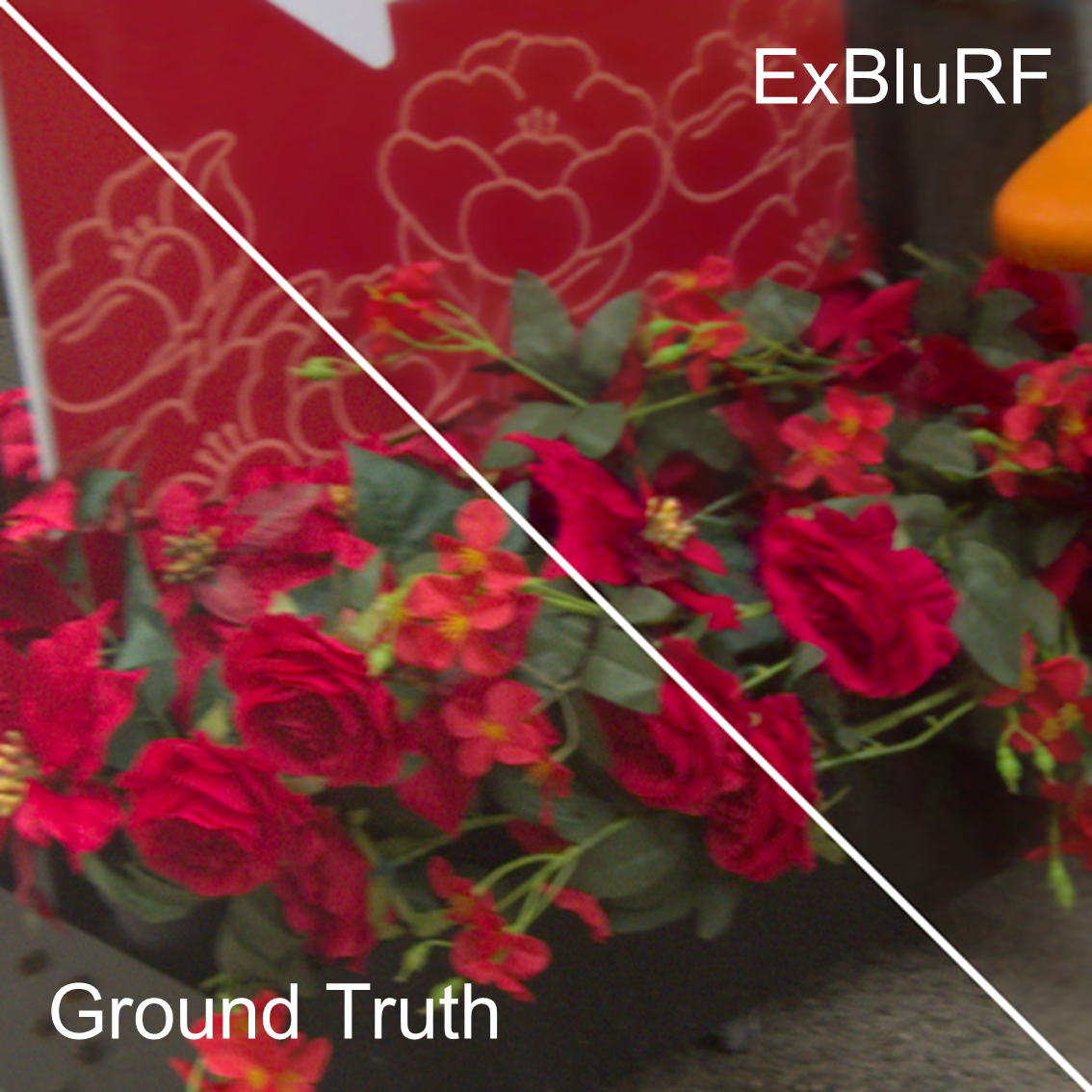}}\hspace*{-0.05in}
	\vspace{-0.05in}
\caption{Given a set of extremely blurred multi-view images (a), our method restores sharp radiance fields and renders clearly deblurred novel views (b).
\vspace{-5mm}
\label{fig:teaser}
}
\end{figure}

One naive solution for this problem is to apply deep learning based 2D image deblurring~\cite{chen2022simple,zamir2022restormer} to the input images before optimizing NeRF.
However, this straightforward approach has two limitations: 
1) Pre-training and fine-tuning strategy of the deep neural network is invalid, since NeRF is per-scene optimized without pairs of sharp images.
2) Independently deblurred multi-view images may yield inconsistent geometry in 3D space, and cannot be recovered with NeRF's optimization.

\begin{figure}[t] %%% t: top, b: bottom, h: here
\vspace{-5mm}
\begin{center}
\includegraphics[width=1.0\linewidth]{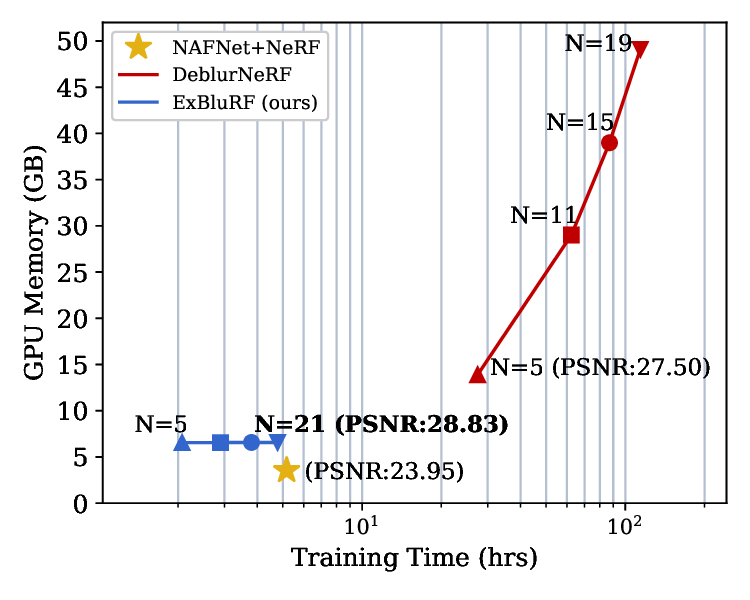}
\end{center}
\vspace{-8mm}
\caption{Training time and GPU memory consumption on ``Camellia" shown in Fig~\ref{fig:teaser}.
Our method, ExBluRF, significantly improves efficiency on both the training time and memory cost with better deblurring performance. $N$ is the number of samples (kernel size) to reconstruct blurry color.}
\vspace{-4mm}
\label{fig:graph_time_vs_gpu}
\end{figure}

The first work that considers the blurry images in NeRF's optimization is DeblurNeRF~\cite{Ma_deblurnerf}.
DeblurNeRF incorporates 2D pixel-wise blur kernel estimation~\cite{campisi2017blind} in front of NeRF's ray marching operation. 
The end-to-end framework of DeblurNeRF restores sharp 3D scenes from moderate motion blurred images.
However, in extreme motion blur cases, the 2D pixel-wise kernel approach is difficult to converge with plausible deblurring, because the blur kernels are spatially varying severely with 6-\textit{degree of freedom} (6-DOF) camera motions.
Also, naive implementation using multi-layer perceptron (MLP) networks becomes a bottleneck of training for the extreme motion blur, since the memory consumption and the computation increase proportionally to the blur kernel size as shown in Fig.~\ref{fig:graph_time_vs_gpu}.

In this paper, we propose a novel and efficient NeRF framework to synthesize sharp novel views from extreme motion blurred images.
Inspired by~\cite{lee2018joint,park2017joint}, our framework formulates the latent blur operation of each image by a trajectory of camera motion that generates blur. 
Like the image capturing of conventional cameras, the colors of rays are accumulated while the origin and the direction of the rays change along the estimated trajectory.
These accumulated colors are optimized to minimize the photo-consistency loss with the colors of the input blurry image.
Each view's trajectory is fully constrained by blur patterns of all the pixels on the image, and the latent sharp 3D scene is optimized to satisfy all multi-view blur observations. 

In addition to the proposed motion blur formulation, we adopt voxel-based radiance fields for efficient GPU memory consumption and training time.
Inherently, the blurry view reconstruction in training time requires explosive computation proportional to the number of sampling along the camera trajectory.
We take advantage of volumetric representation approaches~\cite{chen2022tensorf,fridovich2022plenoxels,muller2022instant,SunSC22,yu2021plenoctrees} that are accelerated by replacing the MLP-based representation to voxel-based representation.
Furthermore, our voxel-based radiance fields keep constant memory to the increasing number of sampling on camera trajectory as shown in Fig.~\ref{fig:graph_time_vs_gpu}.

We conduct extensive experiments to validate the proposed approach on both real and synthetic data with extreme motion blur.
In particular, we propose ExBlur, which provides multi-view blurred images and sequences of sharp images simultaneously captured by a dual-camera system.
The ExBlur dataset enables accurate evaluation of novel view synthesis and optimized camera trajectories on real blurred scenes.
In addition, we analyze the efficiency of the proposed method on memory cost and training speed compared to the previous method~\cite{Ma_deblurnerf}.

To summarize, our contributions are:
\begin{compactitem}
	\item We propose the blur model formulated by 6-DOF camera motion trajectory in NeRF’s volume rendering framework that restores the sharp latent 3D scene without neural networks.
	\item We adopt the voxel-based radiance fields to realize efficient deblurring optimization in terms of memory consumption and training time.
	\item We demonstrate the high-quality deblurring and novel view synthesis of the proposed approach by our ExBlur dataset that presents blurry-sharp multi-view images with ground truth (GT) motion trajectory.
\end{compactitem}

%-------------------------------------------------------------------------

\section{Related Work}
\noindent\textbf{Image Deblurring.}
Deblurring is a long-standing problem in image restoration due to its nature of ill-posedness.
Most conventional methods~\cite{campisi2017blind,cho2009fast,gupta2010single,hyun2014segmentation} estimate a 2D blur kernel that produces a blurred image by convolving on a latent sharp image.
After advance in deep learning and release of large-scale datasets~\cite{nah2019ntire,nah2017deep,rim2022realistic,rim2020real,zhong2020efficient} with blurry-sharp image pairs, recent methods~\cite{chakrabarti2016neural,cho2021rethinking,kupyn2019deblurgan,sun2015learning,tao2018scale,wieschollek2017learning,zhang2020deblurring} are based on supervised learning with convolutional neural networks.

In the novel view synthesis with the NeRF's per-scene optimization, the supervised learning approaches are not straightforward to be applied.
The proposed method follows \cite{lee2018joint,park2017joint} that jointly estimate multi-view sharp images and the blur kernel formulated by a camera motion trajectory and depth maps. Instead of classical optimization~\cite{sun2010secrets} and an auxiliary depth initialization, the optimization of the differentiable volume rendering is employed.

\noindent\textbf{Neural Radiance Fields.}
Neural implicit representation combined with the differentiable volume rendering, NeRF~\cite{mildenhall2020nerf} presents the photo-realistic rendering on arbitrary novel views.
The NeRF-based approaches are applied to numerous 3D vision applications, where human body~\cite{weng2022humannerf}, face~\cite{gafni2021dynamic,park2021nerfies,Sun_2022_CVPR}, hair~\cite{rosu2022neural}, large-scale 3D reconstruction~\cite{hao2021gancraft,turki2022mega}, and simultaneous localization and mapping (SLAM)~\cite{li2022bnv,sucar2021imap,Zhu2022CVPR} are the representative practical applications.

\begin{figure*}
    \vspace{-4mm}
	\centering
	\setlength\tabcolsep{1.5pt}
	\includegraphics[width=\textwidth]{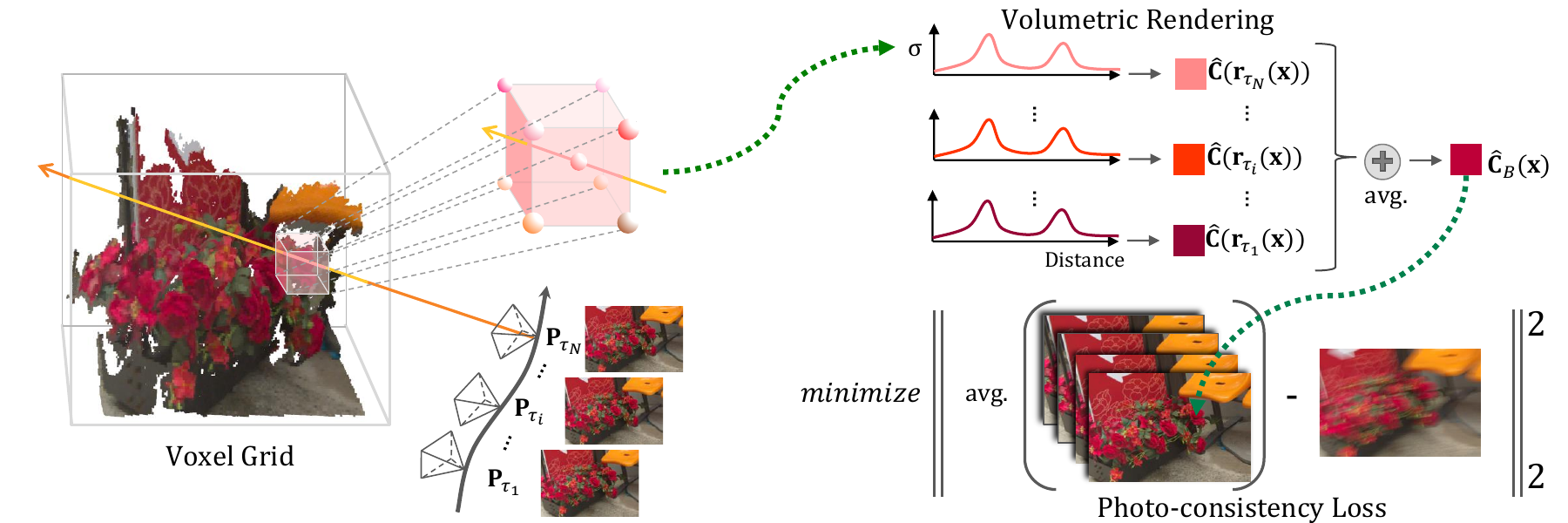}
    \caption{The overview of ExBluRF. We incorporate the physical operation that generates camera motion blur in the volume rendering of radiance fields. The blurry RGB color is reproduced by accumulating the rays along the estimated camera trajectory. By minimizing the photo-consistency loss between the accumulated color and input blurry RGB, we obtain sharp radiance fields and the camera trajectories that explain the motion blur of training views. We adopt voxel-based radiance fields to deal with explosive computation when optimizing extremely motion blurred scenes.}
    \vspace{-4mm}
    \label{fig:pipeline}
\end{figure*}

On the other hand, a number of methods focus on ameliorating the weakness of NeRF for practical usage.
NeRF with degraded multi-view images is constantly studied by expanding the type of image degradation.
Under low-light conditions, noisy and blurry images are acquired inevitably.
NAN~\cite{pearl2022nan} proposes the noise-aware NeRF that deals with burst noised images, and NeRF in the dark~\cite{mildenhall2022nerf} more focuses on HDR view synthesis from noisy images.
For blurry images, DeblurNeRF~\cite{Ma_deblurnerf} introduces the end-to-end volume rendering framework that jointly estimates the pixel-level spatial varying blur kernel and the latent sharp radiance fields.
DeblurNeRF yields plausible deblurring and view synthesis results, while it leaves the scalability to the extreme blur as an open problem.

\noindent\textbf{Voxel-based Radiance Fields.}
While NeRF presents a novel aspect of continuous scene modeling with a neural network, the training and rendering time of the neural implicit representation need to be improved for practical usage.
Many works~\cite{garbin2021fastnerf,lombardi2021mixture,lindell2021autoint,reiser2021kilonerf,wang2022r2l} propose alternative pipelines to NeRF that can reduce the NeRF's training and inference time.
In particular, voxel-based radiance fields approaches~\cite{chen2022tensorf,fridovich2022plenoxels,muller2022instant,SunSC22,yu2021plenoctrees} take advantage of NeRF's differentiable volume rendering, while replacing the MLP-based implicit neural fields to voxel-based representation.
Those methods accelerate the volume rendering by directly interpolating the volume densities and colors from voxel grids.
Implemented in customized CUDA kernels~\cite{cuda}, the voxel-based methods achieve real-time rendering with fast training speed.

%------------------------------------------------------------------------
\section{Method}
We introduce ExBluRF, efficient radiance fields to synthesize sharp novel views from extremely blurred images, and the overall pipeline is shown in Fig.~\ref{fig:pipeline}.

\subsection{Preliminary: NeRF}
The core idea of neural implicit representation~\cite{chen2019learning,mescheder2019occupancy,peng2020convolutional} is to replace the conventional 3D scene representation (\eg point cloud, voxel, mesh) with the per-scene optimized neural network.
The parameters of the network memorize the information for the continuous 3D location $\mathbf{X}\in\mathbb{R}^3$ on the scene. Along with viewing direction $\mathbf{d}\in\mathbb{R}^3$, the neural function of NeRF~\cite{mildenhall2020nerf} $f_{NeRF}$ learns the mapping to the volumetric radiance fields $(\sigma,\mathbf{c})$ as follow:
\begin{equation}
	f_{NeRF}:(\mathbf{X},\mathbf{d})\mapsto(\sigma,\mathbf{c}),
\end{equation}
where $\sigma$ and $\mathbf{c}\in\mathbb{R}^3$ denote a volume density and RGB color, respectively. 
To render the RGB color for a pixel coordinate $\mathbf{x}=(x,y)$ of an arbitrary camera view, NeRF shoots a ray $\mathbf{r}$ into the 3D space and applies the differentiable volume rendering.
Let $\mathbf{P}=[\mathbf{R}, \mathbf{t}]\in \mathbf{SE(3)}$ and $\mathbf{K}\in\mathbb{R}^{3\times3}$ denote the camera extrinsic and intrinsic parameters, respectively. The 3D location and direction of the $k^{th}$ sample point on the ray are defined by:
\begin{equation}
\begin{split}
	\mathbf{X}_k&=\mathbf{t}+s_k\cdot\mathbf{d}, \\
	\text{and}\quad&\mathbf{d}=\mathbf{R}\cdot\mathbf{K}^{-1}[\mathbf{x}]_{+},
	\label{eq:direction}
\end{split}
\end{equation}
where $[\mathbf{x}]_+$ is a homogeneous coordinate of $\mathbf{x}$. 
Along the ray sample points, the color of the ray $\hat{\mathbf{C}}(\mathbf{r})$ is computed by the differentiable volume rendering~\cite{kajiya1984ray} as follows:
\begin{equation}
\begin{split}
	\hat{\mathbf{C}}(\mathbf{r})=\sum^{N}_{k=1}T_k(1-\exp(-\sigma_k\delta_k))\mathbf{c}_k, \\
	\text{where}\quad T_k=\exp(-\sum^{k-1}_{j=1}\sigma_j\delta_j).
	\label{eq:rendering}
\end{split}
\end{equation}
$\delta_k$ is the distance between the $(k-1)^{th}$ and $k^{th}$ sample points.
Note that, NeRF employs the MLP network for the implicit representation $f_{NeRF}$.

\subsection{Motion Blur Formulation}
The proposed method trains sharp radiance fields from blurry multi-view images.
In particular, we focus on camera motion blur, which originated from the camera shake during exposure time.
Physically, the motion blurred RGB image can be interpreted as the accumulated color of rays that hit the pixel while the camera is moving.
Let $\mathbf{r}_\tau(\mathbf{x})$ denotes the ray emitted to the pixel $\mathbf{x}$ at shutter time $\tau$.
The blurred RGB $\hat{\mathbf{C}}_B(\mathbf{x})$ when the shutter is open over $[\tau_o, \tau_c]$ is derived as follow:
\begin{equation}
\begin{split}
	\hat{\mathbf{C}}_B(\mathbf{x}) & =\int^{\tau_c}_{\tau_o}\hat{\mathbf{C}}(\mathbf{r}_{\tau}(\mathbf{x}))d\tau \\
								   & \approx \frac{1}{N}\sum_{i=1}^{N}\hat{\mathbf{C}}(\mathbf{r}_{\tau_i}(\mathbf{x})),
\end{split}
\label{eq:blurformulation}
\end{equation}
where the origin of $\mathbf{r}_{\tau}(\mathbf{x})$ is the 3D location of the camera $\mathbf{t}_\tau$ and the direction of $\mathbf{r}_{\tau}(\mathbf{x})$ is $\mathbf{d}_\tau$ defined in Eq.~\ref{eq:direction}. 
The rays are accumulated continuously while the shutter is open. However, we approximate the integral to a finite sum of $N$ intermediate sub-frames $\tau_i=\tau_o+\frac{i-1}{N-1}(\tau_c-\tau_o)$.

\begin{figure}[t]
        \vspace{-3mm}
	\centering
	\subfloat[]{\includegraphics[height=3.7cm]{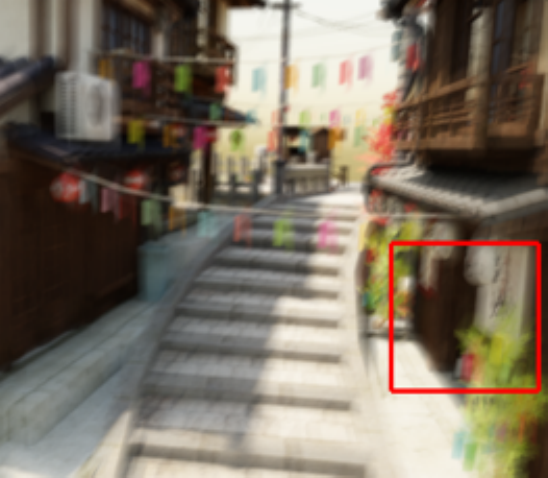}}\hspace*{-0.05in}
        \hfill
	\subfloat[]{\includegraphics[height=3.7cm]{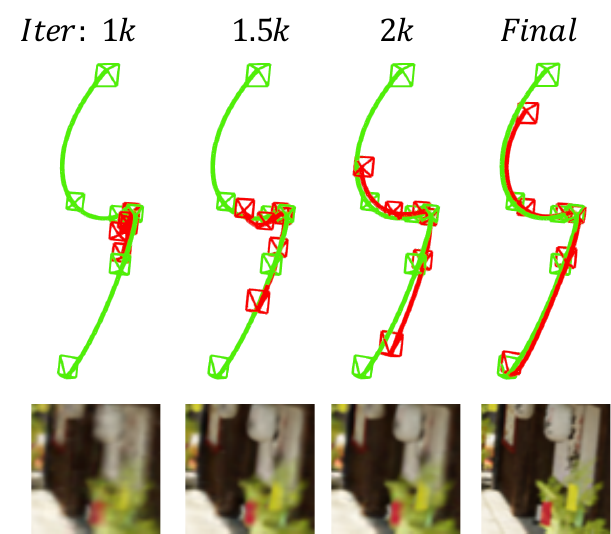}}\hspace*{-0.05in}
	\vspace{-0.05in}
\caption{(a) Blurry training view of synthetic data. (b) Visualization of the GT (\textcolor{green}{green}) and estimated (\textcolor{red}{red}) trajectory of (a) by increasing the training iterations. 
Our blur model converges well to the GT trajectory without sophisticated initialization.
\vspace{-5mm}
}
\label{fig:traj}
\end{figure}

We formulate the blurred RGB $\hat{\mathbf{C}}_B(\mathbf{x})$ by 6-DOF camera poses of intermediated sub-frames. Therefore, the proposed model jointly optimizes the sub-frame camera poses $\{\mathbf{P}_{\tau_i}\}_{i=1}^{N}$ with the radiance fields of the scene. 
Since the camera moves in an arbitrary but smooth trajectory, we re-parameterize the 6-DOF camera motion trajectory by B\'{e}zier curve~\cite{srinivasan2017light}. 
Let $\{\mathbf{p_{\tau_i}}\}^N_{i=1}\in\mathfrak{se}(3)$ denotes Lie algebra of $\{\mathbf{P}_{\tau_i}\}_{i=1}^{N}$, and $\{\hat{\mathbf{p}}_j\}^{M}_{j=0}\in\mathfrak{se}(3)$ is the control points of the $M^{th}$ order B\'{e}zier curve in Lie algebra representation.
The sub-frame camera pose on B\'{e}zier curve is derived as follows:
\begin{equation}
\begin{split}
	\mathbf{p}_{\tau_i}=\sum_{j=0}^{M}&\binom{M}{j}(1-\tau^{\prime}_i)^{M-j}(\tau^{\prime}_i)^j\cdot\hat{\mathbf{p}}_j, \\
	&\text{where} \quad \tau^{\prime}_i=\frac{\tau_i-\tau_o}{\tau_c-\tau_o}.
\end{split}
\end{equation}
$\tau^{\prime}_i$ is normalized time of sub-frame $\tau_i$ over $[0, 1]$.

In contrast to the prior works~\cite{lee2018joint,park2017joint} that model blur formulation using the camera trajectory in light fields or video, we optimize each training view's latent camera trajectory independently.
Also, the proposed model has no prior assumption to initialize the control points $\{\hat{\mathbf{p}}_j\}^{M}_{j=0}$.
However, we found that all the control points initialized with the camera pose estimated using COLMAP~\cite{schonberger2016structure} well converge to the latent camera trajectory as shown in Fig.~\ref{fig:traj}.

Note that, even though we formulate the blur operation in a low-dimensional 6-DOF pose space, the training of MLP-based radiance fields~\cite{mildenhall2020nerf} from extreme camera motion requires heavy computation.
We shoot $N$ rays to synthesize a single blurry pixel with the proposed blur formulation, which increases $\mathcal{O}(N)$ times more memory consumption and training time compared to training radiance fields on sharp images.
The more extreme the camera motion blur, the larger $N$ is required to keep the latent 3D scene sharp.
Therefore, we accelerate the training of the proposed method with memory-efficient voxel-based radiance fields.

\subsection{Voxel-based Radiance Fields}
Following prior works~\cite{chen2022tensorf,fridovich2022plenoxels,yu2021plenoctrees}, we adopt voxel-based radiance fields as an efficient alternative to MLP-based implicit representation of NeRF~\cite{mildenhall2020nerf}.
The advantage of the voxel-based radiance fields is the memory-efficient framework that limits the learnable parameters to the values on its voxel grid. 
While leveraging the differentiable volume rendering of NeRF, we replace the MLP network $f_{NeRF}$ to voxel grid $\mathcal{G}_{\sigma}:\mathbb{R}^3\mapsto\mathbb{R}$ and $\mathcal{G}_{sh}:\mathbb{R}^3\mapsto\mathbb{R}^{9\times3}$, which are output the volume density and degree 2 of spherical harmonic (SH) coefficients for RGB color, respectively.
Let $\mathbf{X}_k$ and $\mathbf{d}$ in Eq.~\ref{eq:direction} denote the 3D location and direction of $k^{th}$ sample point on the ray, the volume density and color of the point are:
\begin{equation}
\begin{split}
	&\sigma_k=\mathcal{G}_\sigma(\mathbf{X}_k), \\
	&\mathbf{c}_k=S(\mathbf{d})^{\intercal}\cdot\mathcal{G}_{sh}(\mathbf{X}_k),
\end{split}
\end{equation}
where $\mathcal{G}_\sigma(\mathbf{X}_k)$ and $\mathcal{G}_{sh}(\mathbf{X}_k)$ are trilinear interpolated from the nearest 8 voxels of $\mathbf{X}_k$, and $S(\mathbf{d}):\mathbb{R}^3\mapsto\mathbb{R}^9$ is SH function that maps viewing direction to view-dependent color.
In contrast to neural radiance fields that require heavy computation of fully-connected layers, the voxel-based radiance fields directly compute the color of a ray from the interpolated values in voxel grids.
By eliminating feed-forward operation, the voxel-based radiance fields achieve real-time volume rendering.
In addition, coarse-to-fine sparse voxel reconstruction combined with a pruning strategy enables memory-efficient training of the radiance fields.
We leverage these strong points of voxel-based radiance fields to optimization of our blurred view reconstruction.

\begin{table*}[ht]
    \vspace{-2mm}
    \centering
    \resizebox{\textwidth}{!}{
	\scalebox{0.2}{
    \begin{tabular}{lccccccccc}
    \toprule
    		 & \multicolumn{3}{c}{Real Motion Blur~\cite{Ma_deblurnerf}} & \multicolumn{3}{c}{Synthetic Motion Blur} & \multicolumn{3}{c}{ExBlur Dataset (Ours)}  \\
    		 Method & PSNR$\uparrow$ & SSIM$\uparrow$ & LPIPS$\downarrow$ & PSNR$\uparrow$ & SSIM$\uparrow$ & LPIPS$\downarrow$ & PSNR$\uparrow$ & SSIM$\uparrow$ & LPIPS$\downarrow$  \\
    \midrule
    Restormer~\cite{zamir2022restormer}+NeRF & 23.22 & 0.682 & 0.289 & 21.15 & 0.541 & 0.419 & 24.33 & 0.571 & 0.545
    \\
    NAFNet~\cite{chen2022simple}+NeRF  & 24.23 & 0.709 & 0.252 & 20.77 & 0.526 & 0.437 & 26.03 & 0.612 & 0.494
    \\
    DeblurNeRF~\cite{Ma_deblurnerf}  & 25.49 & 0.763 & \textbf{0.182} & 23.98 & 0.687 & 0.301 & 28.87 & 0.709 & 0.402
    \\
    BAD-NeRF~\cite{Wang_2023_CVPR} & 22.59 & 0.633 & 0.252 & 21.94 & 0.590 & 0.349 & 27.15 & 0.647 & 0.453 \\
            \textbf{ExBluRF} (Ours)   & \textbf{25.93} & \textbf{0.775} & 0.198 & \textbf{27.81} & \textbf{0.823} & \textbf{0.227} & \textbf{30.17} & \textbf{0.757} & \textbf{0.284}
     \\
    \bottomrule
    \end{tabular}
  	}
  	}
    \caption{Quantitative comparison of novel view synthesis.}
    \vspace{-2mm}
    \label{tab:main}
\end{table*}

\subsection{Loss Functions}
We jointly optimize sharp latent radiance fields and camera trajectories of given blurry training views.
The radiance fields are parameterized by densities and spherical harmonic coefficients of voxels, and the camera trajectories are 6-DOF poses of B\'{e}zier curve's control points.
We compose a blurred color $\hat{\mathbf{C}}_B(\mathbf{x})$ by Eq.~\ref{eq:blurformulation} and minimize the photo-consistency loss $\mathcal{L}_{color}$ with an observed blurry color $C(\textbf{x})$ as follow:
\begin{equation}
    \mathcal{L}_{color} = \sum_{\mathbf{x} \in \mathcal{N}} \norm{\mathbf{C}(\mathbf{x})-\hat{\mathbf{C}}_B(\mathbf{x})}^2_2,
\end{equation}

where $\mathcal{N}$ is the set of training view's pixels in each batch and $\mathcal{L}_{color}$ is minimized with the mean squared error (MSE).
In addition to the color consistency loss, the parameters of voxel grids are regularized by the total variation~\cite{beck2009fast} $\mathcal{L}_{TV}$ and sparsity priors $\mathcal{L}_s$ introduced in \cite{hedman2021baking} as follow:
\begin{equation} 
\begin{split}
	\nonumber
	\mathcal{L}_{TV} &=\sum_{i,k}\sqrt{\partial_x\mathcal{G}_\sigma(\mathbf{v}_{i,k})^2+\partial_y\mathcal{G}_\sigma(\mathbf{v}_{i,k})^2+\partial_z\mathcal{G}_\sigma(\mathbf{v}_{i,k})^2},\\ 
	\mathcal{L}_{s} &= \sum_{k}\log(1+2\sigma_k^2),
\end{split}
\end{equation}
where $\{{\mathbf{v}_{i,k}}\}_{i=1}^{8}$ denotes voxel grids containing $k^{th}$ sample point along the ray.
$\mathcal{L}_{TV}$ guides voxel grids to have smooth volume density while preserving the object's boundaries of the scene, and $\mathcal{L}_s$ leads volume density to sharp distribution along the ray.
Note that the same $\mathcal{L}_{TV}$ is applied to SH coefficients $\mathcal{G}_{sh}(\mathbf{v}_{i,k})$. 
Finally, our combined loss function is set as follows:
\begin{equation}
	\mathcal{L}=\mathcal{L}_{color}+\lambda_{TV}\mathcal{L}_{TV}+\lambda_s\mathcal{L}_s,
\end{equation}
where $\lambda_{TV}$ and $\lambda_s$ are set to $5\times10^{-4}$ and $1\times10^{-12}$ in our experiments, respectively.\\

\subsection{ExBlur Dataset}
To quantitatively evaluate the deblurring performance on real world scenes, we collect a real dataset, named ExBlur dataset, using a dual-camera system similar to ~\cite{rim2020real, rim2022realistic,  zhong2020efficient, zhong2022real}.
The dual-camera system equally splits photons into two cameras using a beam-splitter and simultaneously captures a blurry image and sequence of sharp images, as done in ~\cite{rim2022realistic}.
Specifically, one of the cameras captures a single blurry image with a long exposure time, and the other camera captures a sequence of sharp images during the exposure time of the blurry image.
Using the system, we captured eight scenes, each consisting of 20 to 40 multi-view blurry images and the corresponding sequences of sharp images.
 
We utilize the pairs of blurry and sequence of sharp images in two aspects: 
1) Accurate evaluation of novel view synthesis on real blurred scenes. 
2) Verification of our blur formulation that optimizes latent camera trajectories on real camera shake motion.
The real dataset of DeblurNeRF~\cite{Ma_deblurnerf} provides real blurry images for optimizing NeRF. However, inaccurate camera poses of blurry images lead to misalignments on test views, resulting in erroneous evaluation.
The ExBlur dataset provides accurate camera poses of blurry images by applying structure-from-motion (COLMAP~\cite{schonberger2016structure}) to the corresponding sharp images.
The accurate camera poses confer rigid relative poses to the test views and enable accurate evaluation of novel view synthesis on real blurred scenes.
Furthermore, applying structure-from-motion to other views of sharp sequences produces the ground truth camera trajectory that generates blurry images.
We evidence that our blur optimization is well converged to the ground trajectories by measuring the trajectory evaluation metrics proposed in~\cite{zhang2018tutorial}.

\begin{figure*}[h]
	%\vspace{0.15in}
        %\vspace{-3mm}
	\centering
	\subfloat[Blurry View]{\includegraphics[height=7.25cm]{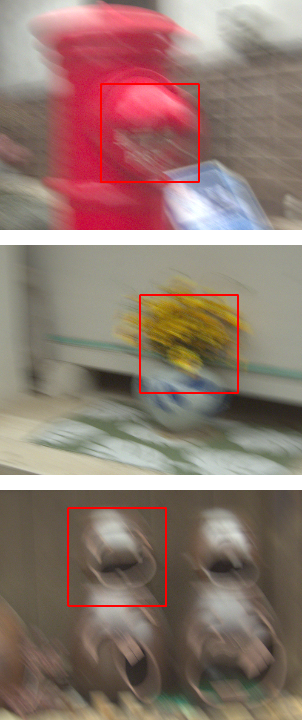}}
        \vspace{-0.05in}
        \hfill
	\subfloat[Input]{\includegraphics[height=7.25cm]{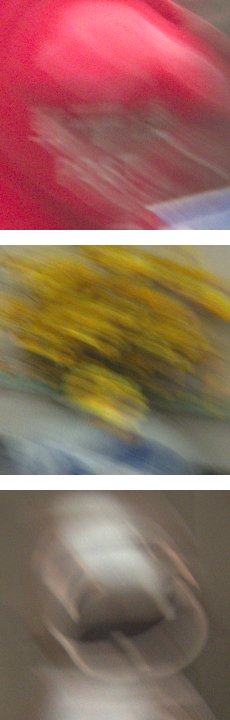}}
	\vspace{-0.05in}
        \hfill
        \subfloat[Ground Truth]{\includegraphics[height=7.25cm]{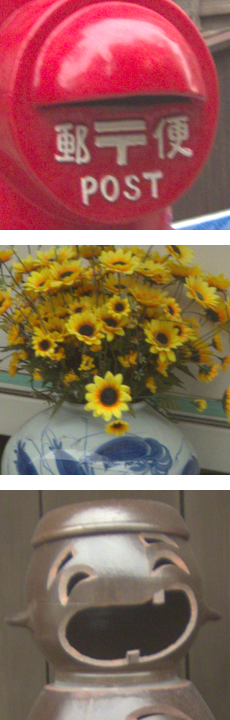}}
	\vspace{-0.05in}
        \hfill
        \subfloat[\cite{chen2022simple}+NeRF]{\includegraphics[height=7.25cm]{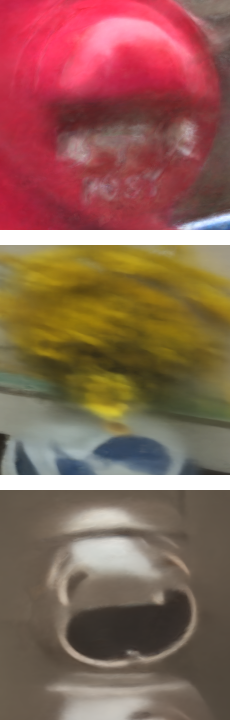}}
	\vspace{-0.05in}
        \hfill
        \subfloat[DeblurNeRF]{\includegraphics[height=7.25cm]{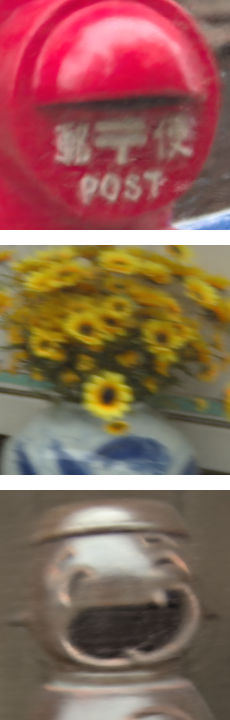}}
	\vspace{-0.05in}
        \hfill
        \subfloat[BAD-NeRF]
        {\includegraphics[height=7.25cm]{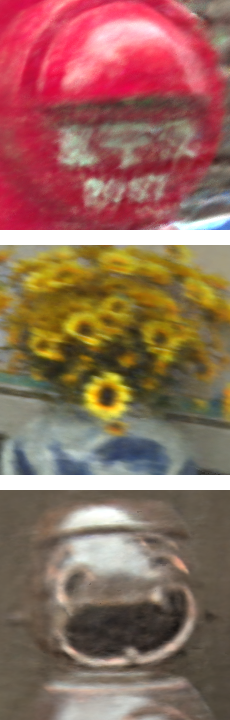}}
	\vspace{-0.05in}
        \hfill
        \subfloat[Ours]{\includegraphics[height=7.25cm]{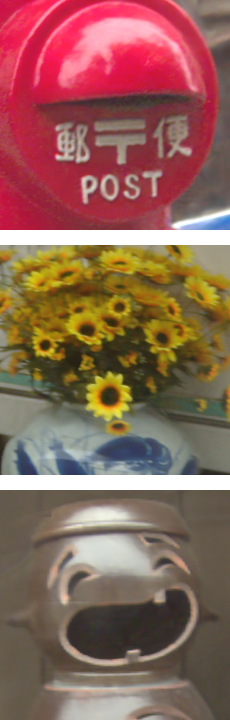}}
	\vspace{-0.05in}
\vspace{8mm}
\caption{Qualitative comparison of deblurring on ExBlur dataset.}
\vspace{-4mm}
\label{fig:exblur}
\end{figure*}

%------------------------------------------------------------------------
\section{Experiments}

\noindent\textbf{Datasets.}
We evaluate our method with three datasets, the real and synthetic datasets from DeblurNeRF~\cite{Ma_deblurnerf} and the proposed ExBlur dataset.
The real dataset of DeblurNeRF~\cite{Ma_deblurnerf} consists of 10 scenes that are captured from a hand-held camera.
Each scene has 20 to 40 blurry training views and 4 to 5 sharp test views.
The camera poses are estimated from blurry training views and sharp test views using COLMAP~\cite{schonberger2016structure}.
For the synthetic dataset, 5 scenes are collected using Blender~\cite{blender}, where each scene consists of 29 blurry training views and 5 sharp test views.
The GT camera poses are exported while rendering using Blender.
In the synthetic dataset of DeblurNeRF~\cite{Ma_deblurnerf}, the synthetic blurry views are generated only with linear motion.
However, for our experiments, we generate more challenging motion blur by random camera motion trajectories in 6-DOF.
The image resolution is $600\times400$ for both the real and synthetic datasets of DeblurNeRF.

For the ExBlur dataset, we collected 8 diverse outdoor scenes with challenging camera motion.
Each scene has 20 to 40 blurry views for training and 4 to 6 test views for evaluation of novel view synthesis.
All blurry images of the ExBlur dataset are paired with sequences of sharp images.
We leverage the sharp images for evaluation of the deblurring performance and accuracy of our estimated camera trajectories.
The resolution for the multi-view images of ExBlur dataset is $800\times540$.

\noindent\textbf{Implementation Details.}
To model complex camera motion with the differentiable curve, B\'{e}zier curve of order $M=7$ is applied and the number of sampling on the trajectory is $N=21$ for the proposed method.
We build our voxel-based radiance fields with a custom CUDA~\cite{cuda} kernel that extends the implementation of ~\cite{fridovich2022plenoxels} to backpropagate gradients to 6-DOF camera trajectory for our blur model.
For the parameters of camera motion trajectory, we utilize Adam~\cite{2015-kingma} optimizer with learning rate $5\times10^{-4}$ and RMSProp~\cite{rmsprop} optimizer is applied for voxel grids.
To optimize radiance fields from blurry images, we train ExBluRF for 200k iterations with a batch size of 25k rays on a single NVIDIA RTX Quadro RTX 8000.
In addition, the pruning and coarse-to-fine strategies are applied for voxel reconstruction, where voxel resolutions of $xy$-dimensions are upsampled every $40k$ iteration.

\begin{figure*}[h]
	\centering
	\subfloat[Blurry View]{\includegraphics[height=7.05cm]{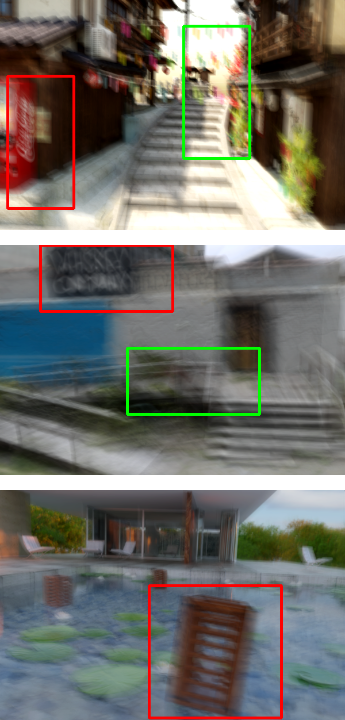}}
        \vspace{-0.05in}
        \hfill
	\subfloat[Input]{\includegraphics[height=7.05cm]{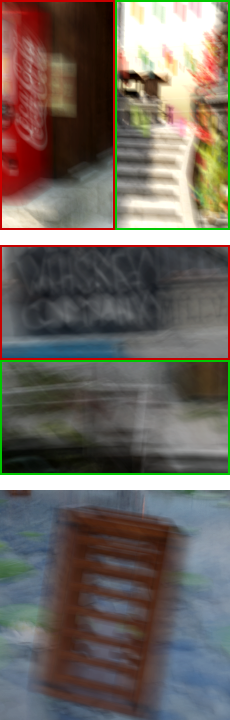}}
	\vspace{-0.05in}
        \hfill
        \subfloat[Ground Truth]{\includegraphics[height=7.05cm]{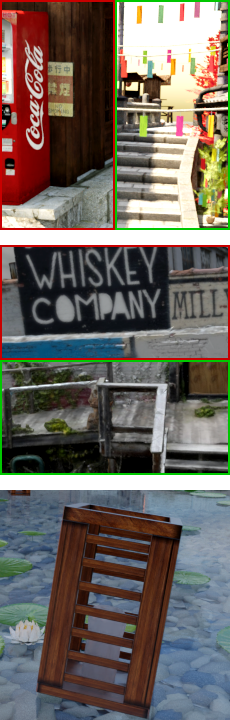}}
	\vspace{-0.05in}
        \hfill
        \subfloat[\cite{chen2022simple}+NeRF]{\includegraphics[height=7.05cm]{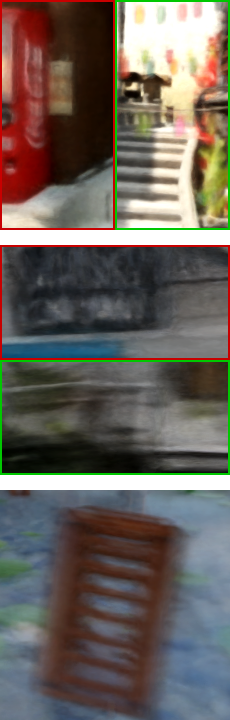}}
	\vspace{-0.05in}
        \hfill
        \subfloat[DeblurNeRF]{\includegraphics[height=7.05cm]{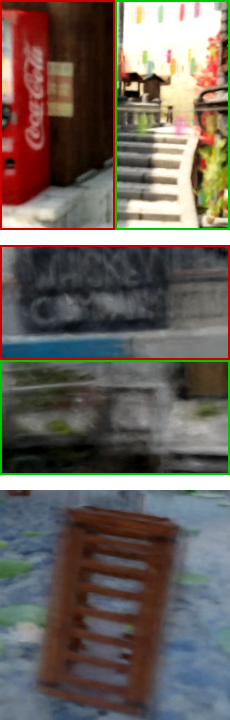}}
	\vspace{-0.05in}
        \hfill
        \subfloat[BAD-NeRF]
        {\includegraphics[height=7.05cm]{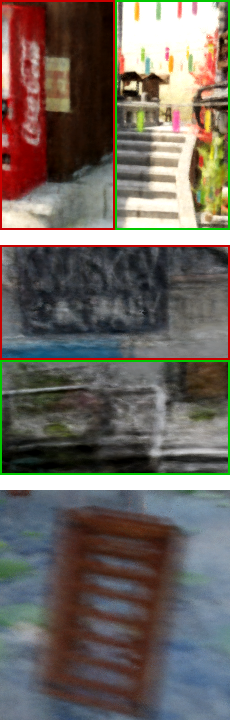}}
	\vspace{-0.05in}
        \hfill
        \subfloat[Ours]{\includegraphics[height=7.05cm]{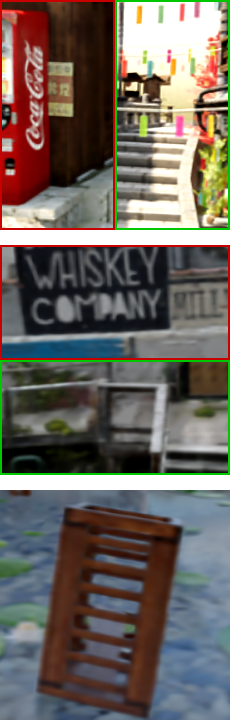}}
	\vspace{-0.05in}
\vspace{8mm}
\caption{Qualitative comparison of deblurring on synthetic dataset.}
\vspace{-5mm}
\label{fig:synthetic}
\end{figure*}

\begin{table*}[ht]

    \vspace{4mm}
    \centering
    \resizebox{\textwidth}{!}{
	\scalebox{0.2}{
    \begin{tabular}{l|c|c|c|c|c|c|c}
    \toprule
    Method & Blur Model & Radiance Fields & Memory cost & $N=5$ & $N=11$ & $N=15$ & $N=19$ \\
    \hline
    DeblurNeRF~\cite{Ma_deblurnerf} & 2D kernel & Neural & $\mathcal{O}(BNPL)+\theta_{NeRF}+\theta_{kernel}$ & 13.89 GB & 29.01 GB & 38.99 GB & 49.09 GB \\
    ExbluRF (Ours) & Camera trajectory & Voxel & $\mathcal{O}(BN)+\phi_{voxel}$ & 6.54 GB & 6.55 GB & 6.56 GB & 6.57 GB \\
        
    \bottomrule
    \end{tabular}
  	}
  	}
    \vspace{-2mm}
    
    \caption{Comparison of memory cost of DeblurNeRF~\cite{Ma_deblurnerf} and our ExBluRF.}
    \vspace{-4mm}
    \label{tab:memory_cost}
\end{table*}

\subsection{Results}\label{results}
We compare the deblurring and novel view synthesis performance with DeblurNeRF~\cite{Ma_deblurnerf}, BAD-NeRF~\cite{Wang_2023_CVPR} and 2D image deblurring methods~\cite{zamir2022restormer,chen2022simple} combined with NeRF~\cite{Ma_deblurnerf}.
DeblurNeRF jointly optimizes neural radiance fields and 2D pixel-wise blur kernel estimation in an end-to-end manner.
BAD-NeRF~\cite{Wang_2023_CVPR} is our concurrent work that models the camera motion blur using the 6-DOF linear camera trajectory.
For our experiments, we follow the default configuration of DeblurNeRF.
To compare with image deblurring methods, we select the two state-of-the-art image deblurring methods that are Restormer~\cite{zamir2022restormer} and NAFNet~\cite{chen2022simple}.
Blurry training views are independently deblurred, and the deblurred images are used to input of vanilla NeRF.\\

\noindent\textbf{Evaluation on Deblurring.}
The rendered novel view images are evaluated by PSNR, SSIM~\cite{hore2010image}, LPIPS~\cite{hore2010image} as metrics.
The three metrics are vulnerable to misalignment, therefore we use camera poses from sharp image pairs for the ExBlur and synthetic datasets for quantitative evaluation.
Since the real dataset of DeblurNeRF~\cite{Ma_deblurnerf} only provides camera poses estimated from blurry images, we re-align the poses of test views by rendered images from radiance fields of each method for quantitative comparison.
After optimization of each method, we render deblurred images of training views with initial poses.
Then, we apply image registration of COLMAP~\cite{schonberger2016structure} to produce aligned poses of test views with deblurred training views.
Note that, the indirect alignment using deblurred results depends on each method's deblurring performance, so the evaluation could not be accurate.

The quantitative results are shown in Tab.~\ref{tab:main}, and qualitative results are shown in Fig.~\ref{fig:exblur} and Fig.~\ref{fig:synthetic}.
Tab.~\ref{tab:main} shows that our method consistently outperforms previous approaches on novel view synthesis.
For Real Motion Blur dataset~\cite{Ma_deblurnerf} that is captured on relatively small motion blur, our method is slightly better than others.
If the camera motion is small, DeblurNeRF with 2D pixel-wise kernel estimation converges to a sharp latent 3D scene as well.
Without considering blur operation in volume rendering, view-dependent deblurring errors of Restormer and NAFNet generate hardly deblurred surfaces when training NeRF, as shown in Fig.~\ref{fig:exblur}.
On the other hand, the proposed method and DeblurNeRF reconstruct 3D consistent scenes, where our approach renders sharper latent 3D scenes.
Note that BAD-NeRF formulates the camera motion in the 6-DOF linear trajectory, and fails to restore the sharp radiance fields from the real-world blurry scenes with the hand-shake camera motion.

In the case of extreme camera motion, a large kernel size is required for DeblurNeRF, otherwise, the latent 3D scene becomes a little blurry not to produce a discontinuity in blurred view reconstruction while training.
The kernel size on DeblurNeRF, the number of sampling on a trajectory in our method, is the key and inevitable factor to deblur extreme motion.
However, DeblurNeRF's naive MLP-based blur kernel estimation network and neural radiance fields demand painfully increasing training time and computation resources.

\begin{table}[t]
	\vspace{2mm}
    \centering
    \setlength{\tabcolsep}{6pt}
    \resizebox{\columnwidth}{!}{

    \begin{tabular}{c|c|c}
    \hline
    		& Notation & Meaning \\
    \hline
    \multirow{3}{*}{Common} & $N$ & Number of sampling for blur operation \\
    & $B$ & Number of ray batch \\
    & $P$ & Number of sample points on a ray \\
    \hline
    \multirow{3}{*}{DeblurNeRF} & $\theta_{kernel}$ & MLP parameters for kernel estimation \\
    & $\theta_{NeRF}$ & MLP parameters for NeRF \\
    & $L$ & The number of MLP layers in NeRF  \\
    \hline
    \multirow{1}{*}{ExBluRF (Ours)} & $\phi_{voxel}$ & Voxel grid parameters \\

    \bottomrule
    \end{tabular}
    }
	\vspace{-2mm}
    \caption{Notations used for the analysis of memory cost.}
    \label{tab:notations}
    \vspace{-4mm}

\end{table}

\subsection{Analysis and Ablation}
\label{sec:Analysis_ablation}
\noindent\textbf{Analysis on Memory Efficiency.}
We examine the efficiency of our model in terms of memory cost as shown in Tab.~\ref{tab:memory_cost}. 
In Tab.~\ref{tab:notations}, we categorize the key factors that influence memory consumption. 
We observe that our method keeps constant memory usage regardless of the number of sampling on a trajectory. On the other hand, DeblurNeRF~\cite{Ma_deblurnerf}'s memory consumption grows proportional to the kernel size.
Both approaches compute total $B \times N$ rays per iteration. 
For our model, each ray marching process sequentially evaluates sample points on the ray from the origin to the far. This only takes $O(1)$ memory space for each ray.  
In contrast, DeblurNeRF~\cite{Ma_deblurnerf} synchronously feed-forwards $P$ points for each ray. 
Moreover, DeblurNeRF~\cite{Ma_deblurnerf}'s network saves activation value from $L$ layers. 
This results in $\mathcal{O}(PL)$ memory cost required for each ray marching.
Consequently, our approach can enjoy a large number of sampling which is a crucial factor of high performance with little memory limitation.
Note that our model has $\phi_{voxel}$ as base model memory which might be larger than DeblurNeRF~\cite{Ma_deblurnerf}'s model size $\theta_{kernel}$ and $\theta_{NeRF}$, while this does not scale by $N$.

\begin{table}[h]
    \centering
    \setlength{\tabcolsep}{6pt}
	\resizebox{\columnwidth}{!}{
    \begin{tabular}{c|c|c|c|c|c}
    \toprule
    	 Blur Model & Radiance Fields & $N$ Sampling & PSNR & GPU Memory  & Training time \\
    \hline
     - & Neural & - & 22.96 & 3.55 GB & 5.17 hrs \\
    2D kernel & Neural & 5 & 27.76 & 13.89 GB & 27.48 hrs  \\
    2D kernel & Voxel & 21 & 27.79 & 6.19 GB & 5.72 hrs \\
    Cam. Traj. & Voxel & 21 & 28.83 & 4.31 GB & 4.41 hrs \\
    \bottomrule
    \end{tabular}
    }
    \caption{Ablation test on the main components in our method. We report "Camellia" scene of ExBlur dataset.}
	\vspace{-2mm}
    \label{tab:ablation_part}
\end{table}

\noindent\textbf{Ablation Study.}
We analyze the effectiveness of the proposed blur model and voxel-based radiance fields in Tab.~\ref{tab:ablation_part}.
The baseline method that consists of an MLP-based 2D kernel estimator and neural radiance fields refers to DeblurNeRF~\cite{Ma_deblurnerf} with a default configuration.
When replacing the neural radiance fields to voxel-based radiance fields with increased kernel size, the computational cost is significantly reduced.
However, the deblurring performance is hardly improved for the extremely motion blurred scene.
The 2D pixel-wise kernel estimation is difficult to leverage the prior knowledge that the camera is moving on a single trajectory and rays are sequentially accumulated.

\begin{table}[h]
	\vspace{-3mm}
    \centering
    \setlength{\tabcolsep}{6pt}
    \resizebox{\columnwidth}{!}{
    \begin{tabular}{l|c|c|c|c}
    \toprule
          \multirow{2}{*}{}  & \multicolumn{2}{c|}{Synthetic} & \multicolumn{2}{c}{ExBlur} \\
    \cline{2-5}
    		& Factory & Tanabata & Postbox & Sunflowers \\
    \hline
        $\text{COLMAP}_{init}$~\cite{schonberger2016structure} & 0.0480 & 0.0422 & 0.0146 & 0.0211 \\
        ExBluRF (Ours) & 0.0141 & 0.0102 & 0.0019 & 0.0040 \\
    \bottomrule
    \end{tabular}
    }
	\vspace{-2mm}
    \caption{Evaluation of absolute trajectory error on ExBlur and synthetic datasets. We report 2 scenes for each dataset.}
    \label{tab:trajectory}
    \vspace{-2mm}
\end{table}

\noindent\textbf{Convergence of camera trajectory.}
To validate estimated camera trajectories from our method, we evaluate absolute trajectory error (ATE)~\cite{zhang2018tutorial} using GT trajectories of the ExBlur and synthetic datasets.
Tab.~\ref{tab:trajectory} shows that our approach minimizes ATE without any additional supervision on the trajectory.
Even though each training view is extremely blurred, and difficult to restore the camera trajectory separately, a single 3D scene is optimized to explain all blurs of training views.
This strong constraint encourages our method to overcome the ill-posedness of deblurring problem and reconstruct sharp radiance fields with accurate camera trajectories.\\

\noindent\textbf{Sensitivity to order of B\'{e}zier curve.}
We investigate how our camera trajectory model is affected by order of B\'{e}zier curve on the ``Pool" scene of the synthetic dataset.
Tab.~\ref{tab:abl_bezier} shows PSNR and ATE metrics on translation and rotation by gradually increasing the number of control points of B\'{e}zier curve.
The performance significantly increases when the order grows from 1 to 3.
The $1^{th}$ order B\'{e}zier curve refers to linear motion in 6-DOF pose space.
For challenging camera motions, the linear motion is difficult to fit the trajectory that generates blur, therefore high-order modeling of motion trajectory is required.
The result of $9^{th}$ order shows the possibility of overfitting when increasing the curve's order unnecessarily. \\

\noindent\textbf{Effectiveness of the number of samplings.}
Tab.~\ref{tab:num_of_sampling} shows the deblurring performance of our method on the ``Postbox" scene of the ExBlur dataset according to the number of sampling points on a trajectory.
As discussed in Sec.~\ref{results}, the number of sampling on a trajectory is crucial to the deblurring performance as more sampling points lead to sharper novel view synthesis.
In particular, the performance of 5 samplings, used in DeblurNeRF~\cite{Ma_deblurnerf}, is significantly degraded compared to the other variants.
The result again shows the limitation of DeblurNeRF~\cite{Ma_deblurnerf}, which cannot handle a large number of sampling points. 
On the other hand, ExBluRF can deal with a large number of sampling points without concern about memory and computation costs.

\begin{table}[t]
    \centering
    \setlength{\tabcolsep}{6pt}
	\resizebox{\columnwidth}{!}{
	\scalebox{0.2}{
    \begin{tabular}{cccc}
    \toprule
    		Order of B\'{e}zier curve & PSNR$\uparrow$ & $\text{ATE}_{pos}\downarrow$ & $\text{ATE}_{rot}\downarrow$ \\
    \midrule
    1 & 25.30 & 0.0942 & 0.367 \\
    3 & 29.65 & 0.0491 & 0.225 \\
    5 & 30.06 & 0.0486 & 0.152 \\
    7 & 30.05 & 0.0454 & 0.163 \\
    9 & 29.92 & 0.0483 & 0.243 \\
    \bottomrule
    \end{tabular}
    }
    }
    \vspace{-2mm}
    \caption{Sensitivity to order of B\'{e}zier curve. We uniformly increase the order by $\{1, 3, 5, 7, 9\}$. The first-order curve means linear motion in 6-DOF pose space.}
	
    \label{tab:abl_bezier}

\end{table}

\begin{table}[t]
	\vspace{-2mm}
    \centering
    \setlength{\tabcolsep}{6pt}
	\resizebox{\columnwidth}{!}{
	\scalebox{0.2}{
    \begin{tabular}{cccc}
    \toprule
    $N$ of sampling on trajectory & PSNR$\uparrow$ & SSIM$\uparrow$ & LPIPS$\downarrow$ \\
    \midrule
    5 & 28.62 & 0.717 & 0.360 \\
    11 & 30.03 & 0.763 & 0.271 \\
    21 & 30.11 & 0.767 & 0.254 \\
    31 & 30.23 & 0.768 & 0.230 \\
    \bottomrule
    \end{tabular}
    }
    }
    \vspace{-2mm}
    \caption{Effectiveness of the number of sampling on the camera trajectory for deblurring performance.}
 \vspace{-4mm}
 
    \label{tab:num_of_sampling}

\end{table}

%------------------------------------------------------------------------
\section{Conclusion}
In this paper, we proposed ExBluRF, efficient radiance fields and blur formulation that restore a sharp 3D scene from extreme motion blurred images.
We formulate motion blur with a 6-DOF camera trajectory and validate that the proposed blur model is well converged to the ground truth trajectory and produces sharp 3D radiance fields.
This blur model is explicitly parameterized in 6-DOF camera poses and combined with computation-efficient voxel-based radiance fields.
We collect the ExBlur dataset that provides extremely blurred multi-view images with ground truth sharp pairs to quantitatively evaluate the deblurring performance on real world scenes.
ExBluRF outperforms existing deblurring approaches both on real and synthetic datasets, while resolving explosive computation cost of the deblurring operation on radiance fields optimization.

\section{Acknowledgment}
This work was supported in part by the IITP grants [No.2021-0-01343, Artificial Intelligence Graduate School Program (Seoul National University), No. 2021-0-02068, and No.2023-0-00156], the NRF grant [No. 2021M3A9E4080782] funded by the Korea government (MSIT), and the SNU-Naver Hyperscale AI Center.

%------------------------------------------------------------------------

{\small
\bibliographystyle{ieee_fullname}
\bibliography{ExBluRF_camera_ready.bib}
}

\end{document}